\documentclass[journal=jacsat,manuscript=article]{achemso}

\usepackage[utf8]{inputenc}
\usepackage{newunicodechar}
\newunicodechar{́}{\'}
\usepackage[version=3]{mhchem} 
\usepackage{xcolor}
\usepackage[table]{xcolor} 
\definecolor{MutedGreen}{RGB}{220, 240, 220}
\definecolor{MutedRed}{RGB}{250, 220, 220}
\usepackage{booktabs}
\usepackage{caption}
\usepackage{multirow} 
\usepackage{fvextra}
\usepackage{amsfonts}
\DefineVerbatimEnvironment{WrapVerbatim}{Verbatim}
  {breaklines=true, breakanywhere=true}

\usepackage{amsthm}
\theoremstyle{definition}
\newtheorem{definition}{Definition}
\author{Ruben Sharma}
\affiliation{Department of Chemical Engineering and Biotechnology, University of Cambridge, Cambridge, UK}
\email{bars2@cam.ac.uk}
\author{Ross D. King}
\affiliation{Department of Chemical Engineering and Biotechnology, University of Cambridge, Cambridge, UK}
\altaffiliation{Department of Computer Science and Engineering, Chalmers University of Technology, Gothenburg,
Sweden}
\email{rk663@cam.ac.uk}

\title[An \textsf{achemso} demo]
  {Compressing Chemistry Reveals Functional Groups}

\abbreviations{}

\begin{document}








\begin{abstract}
We introduce the first formal large-scale assessment of the utility of traditional chemical functional groups as used in chemical explanations.
Our assessment employs a fundamental principle from computational learning theory: a good explanation of data should also compress the data.
We introduce an unsupervised learning algorithm based on the Minimum Message Length (MML) principle that searches for substructures that compress around three million biologically relevant molecules. 
We demonstrate that the discovered substructures contain most human-curated functional groups as well as novel larger patterns with more specific functions.
We also run our algorithm on 24 specific bioactivity prediction datasets to discover dataset-specific functional groups. Fingerprints constructed from dataset-specific functional groups are shown to significantly outperform other  fingerprint representations, including the MACCS and Morgan fingerprint, when training ridge regression models on bioactivity regression tasks.
\end{abstract}
\section{Introduction}
Functional groups are a human-curated set of molecular substructures which are useful for expressing chemical explanations. 
In organic and biological chemistry molecules are often described as several connected functional groups \cite{clayden2012organic}. 
In particular, explanations of chemical and biochemical activity are often described in terms of functional groups. 
For example, the efficacy of several antibiotic compounds
requires the compound to contain the beta-lactam functional group, 
benzodiazepines are characterized by benzene and diazepine functionalities and NSAIDs contain the propionic acid functional group attached to an aromatic group.\cite{bush2016beta,sanabria2021benzodiazepines,elliott1988propionic}. 
\begin{figure}[h]
    \centering
    \includegraphics[width=0.8\linewidth]{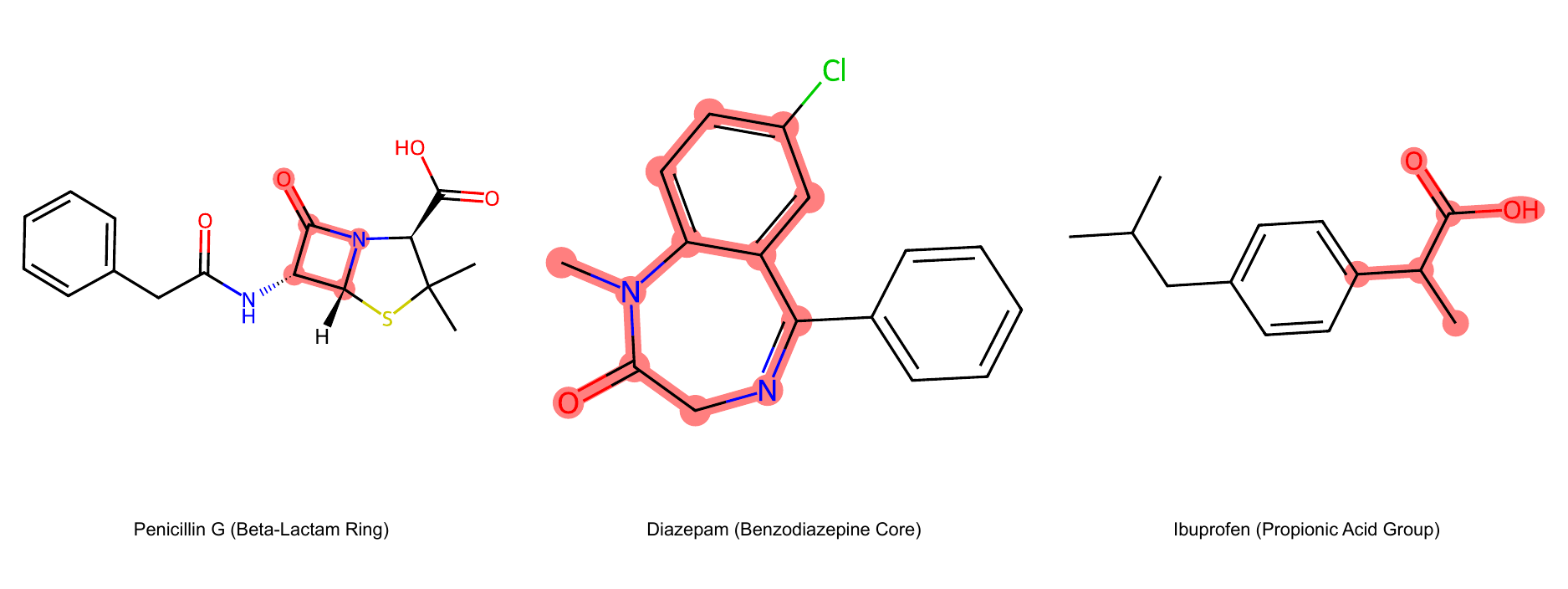}
    \caption{Example drug molecules and necessary functional groups}
    \label{fig:placeholder}
\end{figure}
Caching useful substructures that occur in several explanations enables concise descriptions of molecules and their properties.
Concise explanations are in line with Occam's Razor, a preference for simpler (formalized in computer science as shorter) explanations. 
Specifically, in Solomonoff’s formal theory of inductive inference, the ability of a theory to compress data serves as a formally objective and mathematically optimal measure of the theory’s explanatory and predictive power\cite{solomonoff1964formal}.
Due to their historical utility, functional groups are often used as a  default chemical representation format. In this work we aim to answer the question: 
\begin{enumerate}
\item[] \textit{Do substructures that compress a large corpus of biological molecules correspond to human-curated functional groups?}
\end{enumerate}
There exists no canonical functional group set. A positive answer to our research question could establish an objective basis for constructing such a list.

There are several ways of formulating the compression problem. We choose to employ the Minimum Message Length (MML) principle\cite{wallace2005statistical,allison2018coding}. MML aims to find the most concise explanation of data, where an explanation is a two-part message consisting of a hypothesis (the first part) followed by the data given that the hypothesis is true (the second part). 
Under MML, the most concise explanation trades off the complexity of the hypothesis, and how well the hypothesis fits the data.
A complex hypothesis (large first part) must fit the data very well (small second part) to be selected over a simple hypothesis (small first part) which fits the data moderately well (moderately sized second part).

In this manuscript, we restrict the types of considered hypotheses to `substructure-based explanations'. Specifically, we model the SMILES representation of a chemical dataset as if it were generated by a series of independent draws from a multinomial distribution over SMILES substrings, which correspond to substructures\cite{weininger1988smiles}. We search for the substrings and multinomial probabilities which best compress the dataset according to MML. We use MML because the framework automatically chooses the continuous probability parameters of the multinomial distribution. MML's handling of continuous parameters provides automatic regularisation. 

Specifically, we claim:
\begin{description} 
    \item[\textbf{Claim 1:}] Standard chemical functional groups are objectively useful in general explanations of bioactivity as they emerge during compression of a large corpus of biologically relevant molecules. 
    \item[\textbf{Claim 2:}] Substructures that compress a dataset are useful features for machine learning tasks.
\end{description}
Overall, our contributions are:
\begin{description} 
    \item[\textbf{Contribution 1:}] We introduce an unsupervised algorithm that identifies a set of compressing substructures from a string dataset, in line with the MML principle.
    \item[\textbf{Contribution 2:}] We run our compression algorithm on a dataset containing almost three million biologically relevant molecules represented as SMILES strings and show that the discovered substructures validate conventional functional group theory. We provide a list of these substructures.
    \item[\textbf{Contribution 3:}] We use the substructures learned by our compression algorithm to generate chemical fingerprints. We compare the performance of the chemical fingerprints against MACCS and Morgan fingerprints when learning linear models from bioactivity data\cite{durant2002reoptimization}. 
\end{description}

\section{Related Work}
\textbf{Molecular Representation} Most approaches to molecular representation convert chemical structures into vector representations suitable for computation. Vector representations may be continuous, discrete or propositional. We describe each in turn.

A continuous vector representation of a molecule is a sequence of real numbers
\begin{equation}
    \mathbf{v}_c(m) = [v_1, v_2, \dots, v_n] \in \mathbb{R}^n
\end{equation}
where the vector $\mathbf{v}_c(m)$ encodes information derived from the molecule’s structure, composition, or associated data. 
Examples of continuous vector representations include learned embeddings, where a machine learning model such as a graph neural network learns to map molecular structures into continuous spaces that capture chemical similarity and bioactivity \cite{duvenaud2015convolutional}. In contrast, each vector index in our representation corresponds to a specific feature. 

Discrete vector representations are vectors of integer-valued components. 
A discrete vector representation is defined as
\begin{equation}
    \mathbf{v}_d(m) = [d_1, d_2, \dots, d_n] \in \mathbb{Z}^n
\end{equation}
where each $d_i$ usually encodes a count or categorical indicator of a structural or compositional feature. 
Count-based molecular fingerprints are a common example of discrete vector representations. Each index of a count-based molecular fingerprint corresponds to the integer count of a particular feature, such as a substructure. Binary vector representations are a special case of discrete vector representations, where the components are restricted to elements of the set $\{0,1\}$. Perhaps the most popular discrete vector representation is the Morgan fingerprint \cite{rogers2010extended}. Morgan fingerprints are constructed by enumerating all substructures within a user-defined radius around each heavy atom, assigning each substructure a unique numerical identifier, and hashing these identifiers into a fixed-length binary vector. The hashing operation, however, does not guarantee a one-to-one mapping between substructures and vector indices. Consequently, some indexes of a discrete vector representation may correspond to multiple substructures. Such events are known as `bit collisions' but are rare in practice. 
Our method also represents molecules with a discrete vector representation. In contrast to the Morgan fingerprint, no hashing is applied: each vector index explicitly represents the count of a unique substructure. Moreover, our approach operates on a substantially smaller set of substructures.

Propositional vector representations are vectors with boolean-valued components
\begin{equation}
    \mathbf{v}_p(m) = [p_1, p_2, \dots, p_n] \in \{True, False\}^n
\end{equation}
Each component of a propositional representation corresponds to a logical proposition, which can either be true or false. For example, propositions may correspond to specific abstractions such as `\textit{the chemical has a 7-membered ring}' or `\textit{the chemical contains an actinide}' or specific groups `\textit{the chemical has the functional group} \texttt{NC(O)N}'. One popular propositional representation is the Molecular ACCess Systems keys fingerprint (MACCS) which is of length 166 \cite{rogers2010extended, durant2002reoptimization}. Each of the 166 entries corresponds to the truth value of a proposition. In contrast our approach uses a discrete vector representation, does not consider propositions which do not correspond to fully specified substructures and has variable length depending on the dataset.

\textbf{Compression and Pattern Finding}
The most relevant prior works include the OSCR and MDLCompress algorithms\cite{evans2001symbol, evans2007mdlcompress,evans2007microrna}. Similarly, these algorithms seek to compress the dataset using repeating substrings. In contrast to our approach, these algorithms do not calculate the exact length of the compressed dataset and rely on heuristics. We also modify our algorithm to extract only those substrings which correspond to valid chemical substructures.

\textbf{Identifying Functional Groups}
The most similar work to ours is \citeauthor{erten2019unsupervised} which also uses an Occamist bias to discover functional groups in an unsupervised fashion. 
The authors consider a logic programming approach to functional group learning which is more rigorous, but also computationally much more expensive.
In contrast to our work, the authors do not run the algorithm on a large scale dataset, and do not report all discovered substructures. The authors also do not compare their discovered substructures to existing functional groups, and do not present a method of using them in supervised learning. 
Another similar work is \citeauthor{ertl2017algorithm}. The authors define functional groups as those which conform to a user specified algorithm. Conversely, we do not impose any definition regarding the structure of the groups we wish to extract, except those which are syntactically invalid. We simply aim to find  groups which compress a dataset. 

\section{Algorithms}
To test our claims we require algorithms which (i) identify substructures which compress chemical datasets and (ii) use these compressing substructures in a chemical representation. Our contribution in this section is the introduction of two such algorithms: \textsc{FGCompress} and \textsc{FGFingerprinter}.
\subsubsection{\textsc{FGCompress}}
We now describe our substructure discovery algorithm \textsc{FGCompress}. Our algorithm is a greedy search, choosing substructures which best compress the dataset at each iteration. The algorithm terminates when no substructure can further compress the dataset. 
The algorithm requires a set of molecules represented as SMILES strings as input. A description of the \textsc{FGCompress} procedure follows:
\begin{enumerate}
    \item Enumerate all substrings up to a user-specified maximum length in the dataset.
    \item Filter out substrings that do not correspond to valid chemical substructures. The remaining substrings are termed valid substrings.
    \item For each valid substring, compute the total message length obtained if:
    \begin{enumerate}
        \item All instances of the substring in the dataset are replaced by a single new symbol,
        \item The new symbol–substring pair is added to the codebook, and
        \item the updated codebook and dataset are transmitted as a message.
    \end{enumerate}
    \item Select the substring that yields the shortest total message length (maximally compresses the dataset).
    \item If the selected substring reduces the total message length, add it to the codebook and return to step 1.
    \item Otherwise, if no new substring can further reduce the message length, terminate and return the current codebook.
\end{enumerate}
If a newly added substring (step 4) contains an existing codebook entry, the count of that entry is reduced accordingly. If the count of an entry reaches zero, the substring is removed from the codebook.
A diagrammatic representation of a hypothetical \textsc{FGCompress} run is shown in Figure \ref{fig: Theories} below.
\begin{figure}[H]
\centering
\begin{tabular}{cccc}
Step    &   Codebook &  Compressed Data & Length/bits\\ \hline
\rowcolor{MutedGreen}0   & \texttt{\{\}} & \texttt{\{C=C(Br)C=O, C=CCOC=O, CC=CC=O\}} & 32.1\\ 
\rowcolor{MutedGreen}1   & \texttt{\{X:C=\}} & \texttt{\{XC(Br)XO, XCCOXO, CXXO\}} & 18.2\\ 
\rowcolor{MutedGreen}2   & \texttt{\{Y:C=C, X:C=\}} & \texttt{\{Y(Br)XO, YCOXO, CXXO\}} & 16.6\\ 
\rowcolor{MutedRed}3   & \texttt{\{Z:C=O, Y:C=C, X:C=\}} & \texttt{\{Y(Br)Z, YCOZ, CXZ\}} &17.0\\ 
\end{tabular}
\caption{An illustration of an example run of the \textsc{FGCompress} algorithm. At iteration 2, adding the best substring \texttt{C=O} does not further compress the message. The algorithm is then terminated and the codebook at step 2 is taken as final. Length values are invented for illustrative purposes.}
\label{fig: Theories}
\end{figure}
At each step, the message length is calculated as the number of bits required to transmit the codebook and dataset, modelling molecules as draws from a multinomial distribution specified by the codebook.
We describe specific calculation of the message length in the next section.

\subsubsection{\textsc{FGFingerprinter}} This algorithm takes as input the set of compressing substructures from \textsc{FGCompress} and generates a molecular fingerprint. The fingerprint is a fixed length integer vector of counts. Each index of the vector corresponds to a substructure which was discovered during the \textsc{FGCompress} search. To generate a molecular fingerprint for a molecule, \textsc{FGFingerprinter} counts the number of times each substructure occurs in the molecule and sets the corresponding vector index to this count value. We refer to this fingerprint as the \textsc{MML87} fingerprint, as the \textsc{FGCompress} search procedure relies on the MML87 approximation to calculating the MML codelength\cite{wallace1987estimation}. The MML87 fingerprint is a lossy chemical representation as it does not retain information pertaining to connections between the substructures. 
\section{Definitions}
\begin{definition}[\textbf{SMILES symbol}]
\label{def:SMILES-symbols}
A SMILES symbol is an element of the SMILES alphabet as defined in \citet{weininger1988smiles}. We denote the SMILES alphabet as $\Sigma$.
\end{definition}
\begin{definition}[\textbf{Valid Substring}]
\label{def:Valid Substring}
Several concatenations of SMILES symbols do not correspond to valid substructures. To ensure our codebook contains as few invalid substructures as possible, we filter the set of all substrings. 

The set of all substrings $\Sigma^*$ is the set of all finite strings which can be constructed from $\Sigma$. The set of valid substrings $S=F(\Sigma^*) \subset \Sigma^*$  is the set of all strings that satisfy the filter function $F : \Sigma^* \rightarrow \Sigma^*$. A valid substring is an element of the set of valid substrings $s \in S$. In this report, $F$ filters $\Sigma^*$ for those substrings using the following rules:
\begin{itemize}
    \item If a substring contains a bracket, then it must also contain the corresponding matching bracket
    \item If the substring contains a bond character (=,\#,-,/,\textbackslash) or stereogenic center character (@), then it must contain at least one atom connected to this character.
    \item If the substring contains a number, then it must also contain a second instance of that number (c1cccc not permitted).
    \item The . character is not permitted in any substring.
\end{itemize}
\end{definition}
\begin{definition}[\textbf{Codebook}]
A codebook is a set of pairs $\{(s_1,p_1), ...(s_n,p_n)$, where $s_i \in S$ is a valid substring  and $p_i \in [0,1]$ is a probability value. The probabilities are constrained such that $\sum_{i=1}^{n} p_i = 1$. The codebook is interpreted as a multinomial probability distribution: substring $s_i$ has probability $p_i$ of being drawn randomly, with replacement. 
\end{definition}
\begin{definition}[\textbf{Optimal Coding Scheme}]
\label{def:optimal coding scheme}
We represent the codebook and compressed dataset as binary strings. To ensure our encoding is as concise as possible, we use an optimal coding scheme to represent each part of the message\cite{shannon1948mathematical, cover1999elements}. 
An optimal coding scheme is designed to minimise the average length of an encoded message. The scheme assigns shorter binary strings to more probable events and long strings to less probable ones.  
Formally, an event $x$ with probability $P(x)$ is represented as a binary string of length $-\log_2{P(x)}$. Hereon, all $\log$ terms are assumed to be base 2.
An optimal coding scheme ensures that the average length of the encoded data approaches the entropy of the true symbol generating distribution, the theoretical limit of compression.
An optimal code is a code from an optimal coding scheme.
\end{definition}

\section{The Message}
\begin{figure}
    \centering
    \includegraphics[width=0.3\linewidth]{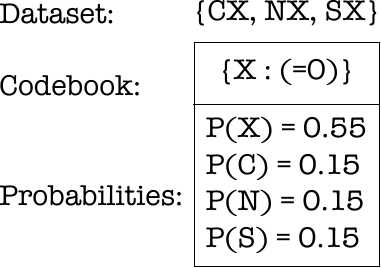}
    \caption{Example Message Contents for the Original Dataset $\texttt{\{C(=O), N(=O), S(=O)\}}$}
    \label{fig:example message contents}
\end{figure}

We now describe the type of message that we send. The length of the message is used as our measure of compression and guides the search towards maximally compressing substrings. We use the terms codebook, valid substring and optimal coding scheme as defined in the previous section. Before the message is communicated, we assume that the sender and receiver have agreed that the probability of a SMILES symbol occurring in a codebook substring is equal to its relative frequency in the dataset. 
Consequently, the receiver is assumed to already know the symbol frequencies in the dataset.
This assumption is equivalent to assuming that the receiver is aware of some preliminary analysis on the dataset but does not know of a good codebook that compresses the data further. The message therefore consists of this codebook, and the dataset compressed by the codebook. An example illustration of the message contents is shown in Figure \ref{fig:example message contents}.
\section{Message Length Calculation}
We now describe how we calculate the message length, the quantity we seek to minimise during our search procedure. Conventionally, a message is described in two parts: sending the codebook (and probabilities), and sending the dataset.
Instead, for clarity we describe the message in three parts. Part 1 communicates the substrings in the codebook. Part 2 sends both the probability values in the codebook and the substring and symbol counts in the dataset given part 1. Part 3 communicates the specific sequence of symbols and substrings given parts 1 and 2. We split the message into these three parts because there exists a known expression for part 2 that we use in our implementation. 
We use the term `vocabulary' to refer to the minimal set of substrings and symbols required to cover the full dataset. In Figure \ref{fig:example message contents}, the vocabulary is $\{\texttt{X,C,N,S}\}$.

The length of the message communicating part 1 is determined by sending each part piecewise. As discussed, the length of communicating part 2 has an already known closed form. The length of communicating part 3 is calculated from the multinomial coefficient: the number of sequences one can make from a dataset of length $N$, composed of $i$ symbols with frequencies $M_1, M_2,...M_i$, where all sequences are equally likely.
We now describe calculation of the three parts in order.

\subsection{Part 1: Communicating the substrings}
First the sender sends each substring in the codebook.
The message that specifies all substrings is structured hierarchically. 
The sender first specifies the total number of substrings in the codebook, $|H_s|$ and then sends the data for each substring sequentially. The message format is

\texttt{[Total Substring Count] [Substring$_1$] [Substring$_2$]\dots}

\noindent Each individual substring is also a two-part message, specifying its length followed by the constituent symbols. The individual substring message is then:

\noindent \texttt{[Substring$_i$]=[Substring$_i$:length][Substring$_i$:Symbol 1][Substring$_i$:Symbol 2] \dots}

\noindent All integer values (total substring count and individual substring lengths) are communicated using an integer code. 
The substring symbols are encoded using their probabilities. 
\begin{enumerate}
    \item \textbf{Encoding integer values:} All integers $N$ are encoded using the logstar universal code\cite{rissanen1983universal}. This is a special code which allows for specifying any integer in a concise manner. Under the logstar code, the transmission cost for the integer is $\log^*(N) \approx \log(N)+\log\log(N) + ...$ bits.
    \item \textbf{Encoding substring symbols:} Each symbol $s$ in substring $S_i$ is encoded according to their pre-agreed probabilities of occurrence, as discussed at the beginning of this section. Under an optimal coding scheme, the cost of transmitting symbol $s$ is $-\log_2{P(s)}$ bits.
\end{enumerate}
The cost of specifying a single substring $C_{string}(S_i)$ is the sum of the cost of its length and costs of all constituent symbols (equation \ref{eq:single substring cost}).
\begin{equation}
    C_{string}(S_i) = \log^*{|S_i|}-\sum_{s\in\Sigma}\left[\text{count}(s,S_i)\times\log{P(s)}\right]
    \label{eq:single substring cost}
\end{equation}
Where $\text{count}(s,S_i)$ is the number of times symbol $s$ appears in substring $S_i$.

The total cost of all of the substrings is the cost of specifying the number of total substrings, plus the sum of the costs of each substring (equation \ref{eq:total substring cost}).
\begin{equation}
    P_1 = \log^*{|H_s|} + \sum_{i=1}^{|H_s|}{C_{string}(S_i)}
    \label{eq:total substring cost}
\end{equation}
For example, consider sending the example substring $S_{ex} = \texttt{CCCN}$ where $P(\texttt{C})=0.5$ and $P(\texttt{N})=0.25$.
\begin{itemize}
    \item The cost to transmit the length ($|S_{ex}|=4$) is $\log^*(4)\approx 2$ bits.
    \item The cost of transmitting the three \texttt{C} symbols is $3\times(-\log{0.5}) = 3$ bits.
    \item The cost of transmitting the single \texttt{N} symbol is $1\times(-\log{0.25}) = 2$ bits.
\end{itemize}
The total length $C(S_{ex})$ is then $2 + 3 + 2 = 7$ bits.
\subsection{Part 2: Sending Vocabulary Probabilities and Counts}
The part 2 message communicates two pieces of information (i) the underlying probabilities assigned to each item in the vocabulary and (ii) the specific counts of those items observed in the dataset. A known approximation to the optimal length of this combined message is given by the MML87 formula for the multinomial distribution. We state this here and refer the reader to \citet{wallace2005statistical} and \citet{wallace1987estimation} for a full derivation.
The length of this part of the message, denoted $P_2$ is
\begin{equation}
P_2 = \log_2 \frac{\Gamma(N + M)}{\Gamma(M)} + \sum_{m} \log_2 \frac{1}{\Gamma(s_m + 1)} + \frac{1}{2} \log((M - 1)\pi) - 0.4
\end{equation}
Where $N$ is the total number of substrings and symbols in the dataset, $M$ is the number of distinct substring and symbols, $\Gamma(.)$ is the gamma function, $s_m$ is the count of the $m^{th}$ symbol and $\pi$ is the standard mathematical constant. 
\subsection{Part 3: Specifying the Sequence of the Symbols}
While the part 2 submessage specifies the number of symbols and their probabilities, it does not communicate their ordering. The final part of the message encodes the specific sequence of the $N$ total substrings and symbols in the encoded dataset. The number of unique sequences that can be formed is given by the multinomial coefficient. Assuming all valid orderings are equally likely, then the probability of a single sequence is simply $\frac{1}{\text{number of unique sequences}}$. Under an optimal code, the cost of specifying the specific sequence the negative logarithm of the probability of a specific sequence, denoted $P_3$. We assume that the receiver knows the number of individual symbols in each SMILES string, so that they can split the received string into the original molecules. 
\begin{equation}
P_3 = \log_2 \frac{N!}{s_1! s_2! \cdots s_M!}
\end{equation}
\subsection{Full Message Length}
The full message length is the sum of the lengths of the three parts
\begin{equation}
    M = P_1 + P_2 + P_3
    \label{eq:full equation}
\end{equation}
We use equation \ref{eq:full equation} as our cost function, and implement a search procedure to minimise this value.
\section{Experiments}
We conducted experiments to verify our claims. We first try to answer the question:
\begin{enumerate}
\item[\textbf{Q1}] Do functional groups emerge when compressing a large corpus of biologically relevant molecules?
\end{enumerate}
We answer \textbf{Q1} by running the \textsc{FGCompress} search procedure on all molecules in the ChEMBL dataset for 500 iterations, and caching the discovered substructures. We terminate the algorithm after 500 iterations due to computational constraints. While early termination may prevent the discovery of additional patterns, \textsc{FGCompress} extracts substructures in order of importance, so the top 500 substructures are expected to strongly represent the dataset.
Unfortunately, there is no canonical list of functional groups to which we could compute a similarity score. 
Instead, we manually assess the discovered substructures, and note exact and partial matches to human, named functional groups. 

To test our claim that the discovered substructures aid learning, we try to answer the question:
\begin{enumerate}
\item[\textbf{Q2}] Are the substructures extracted from compressing the dataset useful for learning?
\end{enumerate}
We answer \textbf{Q2} by generating a unique MML87 fingerprint for each of 24 bioactivity prediction datasets using the \textsc{FGFingerprinter} algorithm.
We compare the predictive performance of three fingerprint representations: our MML87 fingerprint, the 166-bit MACCS fingerprint, and a Morgan fingerprint (radius 2) hashed to the same length as the corresponding MML87 fingerprint for that specific dataset\cite{durant2002reoptimization, rogers2010extended}.
We train a cross-validated ridge regression model for each dataset\cite{stone1974cross, hoerl1970ridge}. We compare the mean squared error of the models on a test set using each fingerprint.

\subsection{Experimental Choices}

We make the following experimental choices.

\textbf{\textsc{FGCompress}}  
We ran the algorithm on the ChEMBL dataset (release 36), which contains 2,878,135 bioactive molecules \cite{gaulton2012ChEMBL}.  
We used a maximum substring enumeration length of 8 and a uniform prior distribution over all substring probabilities via a generalized beta function.   

\textbf{\textsc{FGFingerprinter}}  
We ran the algorithm on 24 bioactivity datasets as reported in \citet{cortes2018deep}, generating 24 unique fingerprints.  
For these smaller datasets, we used a maximum substring enumeration length of 15. Duplicate compounds indistinguishable by SMILES strings were removed using RDKit \cite{landrum2013rdkit}.

\textbf{Models and Hardware}
We ran the \textsc{FGCompress} experiments on a server with 36 CPU cores and 125 GB RAM.
We ran the \textsc{FGFingerprinter} experiments on an M2 MacBook Air (8 CPU cores, 16 GB RAM). All experiments were repeated 5 times, and we report the mean and standard error in each case.  
Ridge regression models were implemented using scikit-learn \cite{pedregosa2011scikit}. Each dataset was split into 75\% training and 25\% testing. The ridge $\alpha$ hyperparameter was optimized via leave-one-out cross-validation on the training set, selecting from $\{0.001, 0.01, 0.1, 1\}$.  

We choose ridge regression models over more expressive models such as random forests, gradient-boosted trees, or neural networks because our goal is to directly evaluate the quality of the substructures. Random forests and gradient-boosted trees implement learned feature transformations on top of the input features, while neural networks construct internal representations in their hidden layers.\cite{breiman2001random, he2014practical, Goodfellow-et-al-2016} In contrast, ridge regression provides a purely linear mapping from features to predictions, making it ideally suited for assessing raw representation quality.

\textbf{Significance Tests}
Statistical significance was assessed using the Wilcoxon signed-rank test with Benjamini-Hochberg multiple testing corrections at $\alpha = 0.05$ \cite{wilcoxon1992individual, benjamini1995controlling}.

\section{Results and Discussion}
We now present the empirical results to address our experimental questions \textbf{Q1} and \textbf{Q2}.
\subsubsection{Q1: Functional group verification}
The full list of substructures discovered by \textsc{FGCompress} may be found in the appendix. In all figures, substructures are listed in order of the iteration at which they were derived. This reflects the explanatory power of the substructure with respect to the dataset. The resulting substructures compress the dataset by 30\% relative to the dataset encoded in terms of SMILES symbols only.

Of the 500 substructures, 494 were successfully converted to SMARTS patterns using RDKit. A substructure was considered convertible if it is already a valid SMARTS pattern or becomes valid by adding the wildcard `\texttt{*}' at the beginning, end, or both. Consequently, we conclude that use of the SMILES representation in our procedure did not significantly affect the interpretability of the extracted substrings.

After filtering for substrings that can be converted to valid SMILES patterns, we further remove substructures that are identical up to numerical relabelling. For example, the substring \texttt{c2ccccc2} would be removed if \texttt{c1ccccc1} is already present. After filtering, \textsc{FGCompress} returned 304 substructures. We next analyze the filtered list.

The top 15 substructures found by the algorithm are shown in Figure \ref{fig:Top15} below. All of these substructures are well known functional groups. For example, the top five groups include the carbonyl, trifluoromethyl, methyl, amide, benzene functional groups. As ChEMBL is a bioactivity dataset we expect functional groups that are strongly represented in the biological literature, such as those which are derivatives of amino acids (substructures 4, 6, 9, 11, 12, 14). Substructures 4 and 9 are identical as they are two SMILES representations of the amide group \texttt{C(=O)N} and \texttt{NC(=O)} that were not caught by our filtering procedure. The duplication of the amide group is a direct artifact of the SMILES representation.

\begin{figure}[H]
\centering
\includegraphics[width=.9\linewidth]{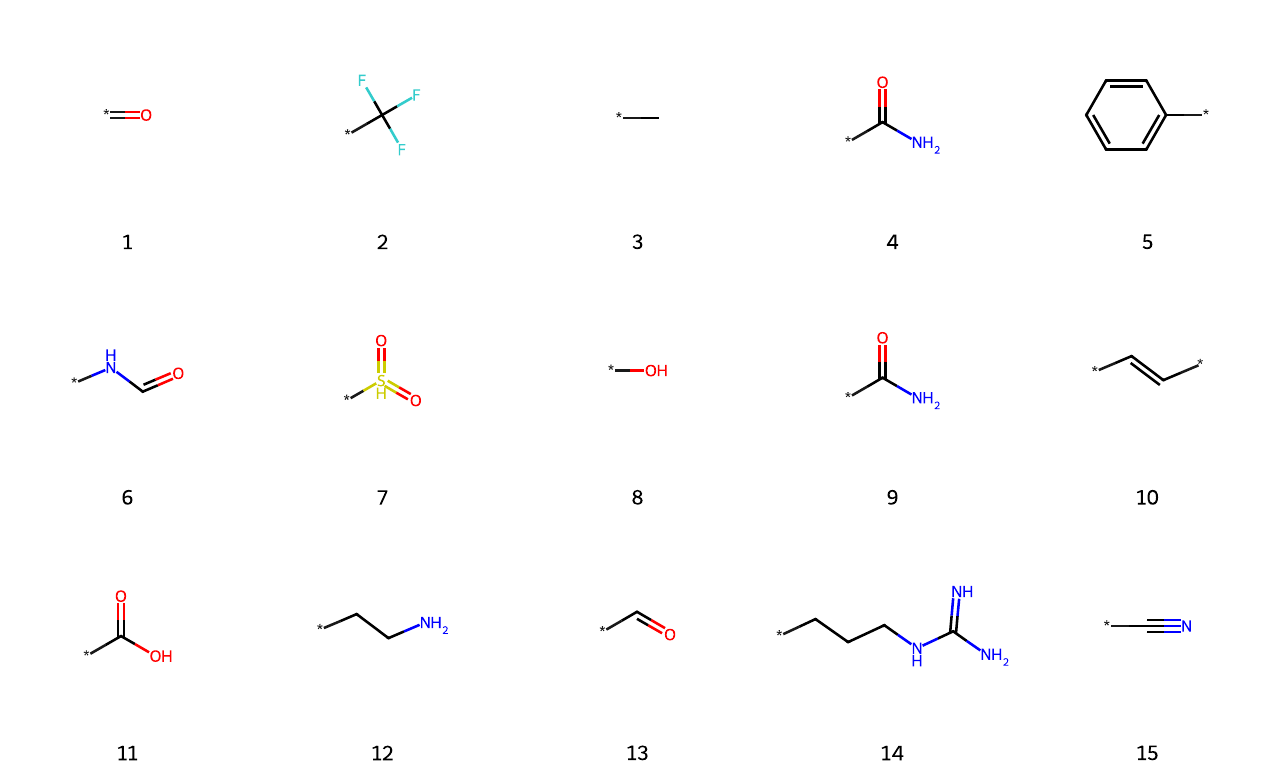}
\caption{Top 15 substructures derived by the algorithm. Substructures are numbered in order of relative importance.}
\label{fig:Top15}
\end{figure} 

We now analyse subsets of the discovered groups: branches, small and large rings. 

\begin{figure}[H]
\centering
\includegraphics[width=1\linewidth]{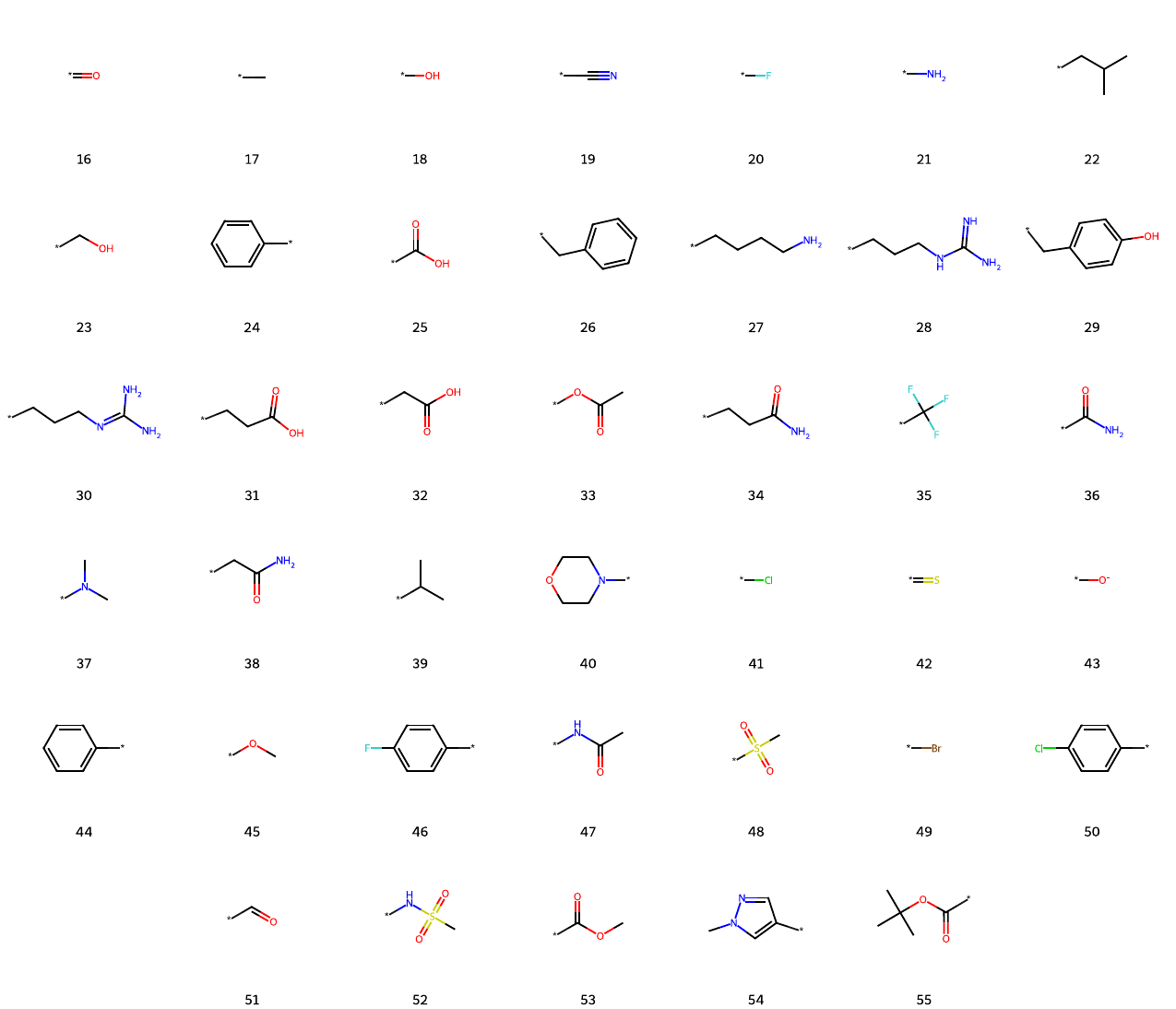}
\caption{Top 40 substructures extracted as complete branches discovered by \textsc{FGCompress}. Substructures are numbered in order of relative importance}
\label{fig:branches}
\end{figure} 
The top 40 branch substructures found by the algorithm are shown and numbered in Figure  \ref{fig:branches}. Nearly all discovered branches are standard human functional groups. For example, the branches include carbonyl (16),  methyl (17), alcohol (18) nitrile (19), fluoro (20) and amine (21) functional groups. As noted previously, due to the nature of the ChEMBL dataset the number of peptide related functional groups is high. Specifically, there is an abundance of amino acid side chains and derivatives (substructures 22, 23, 26, 27, 28, 29, 30, 31, 32, 34, 38, 39). 
Other standard functional groups include benzyl (24, 44), trifluoromethyl (35), tertiary amines (37), morpholine (40), aromatic halides (46, 50) and the Boc protecting group (55).  
Interestingly, all halide substituents in the top 40 branches are in the para configuration. We were unaware a priori of the relative prevalence of ortho, meta, or para configurations.

The top 35 small ring containing substructures that may be expressed in fewer than 20 SMILES symbols are presented in Figure \ref{fig:small ring groups} below.
\begin{figure}[H]
\centering
   \includegraphics[width=1\linewidth]{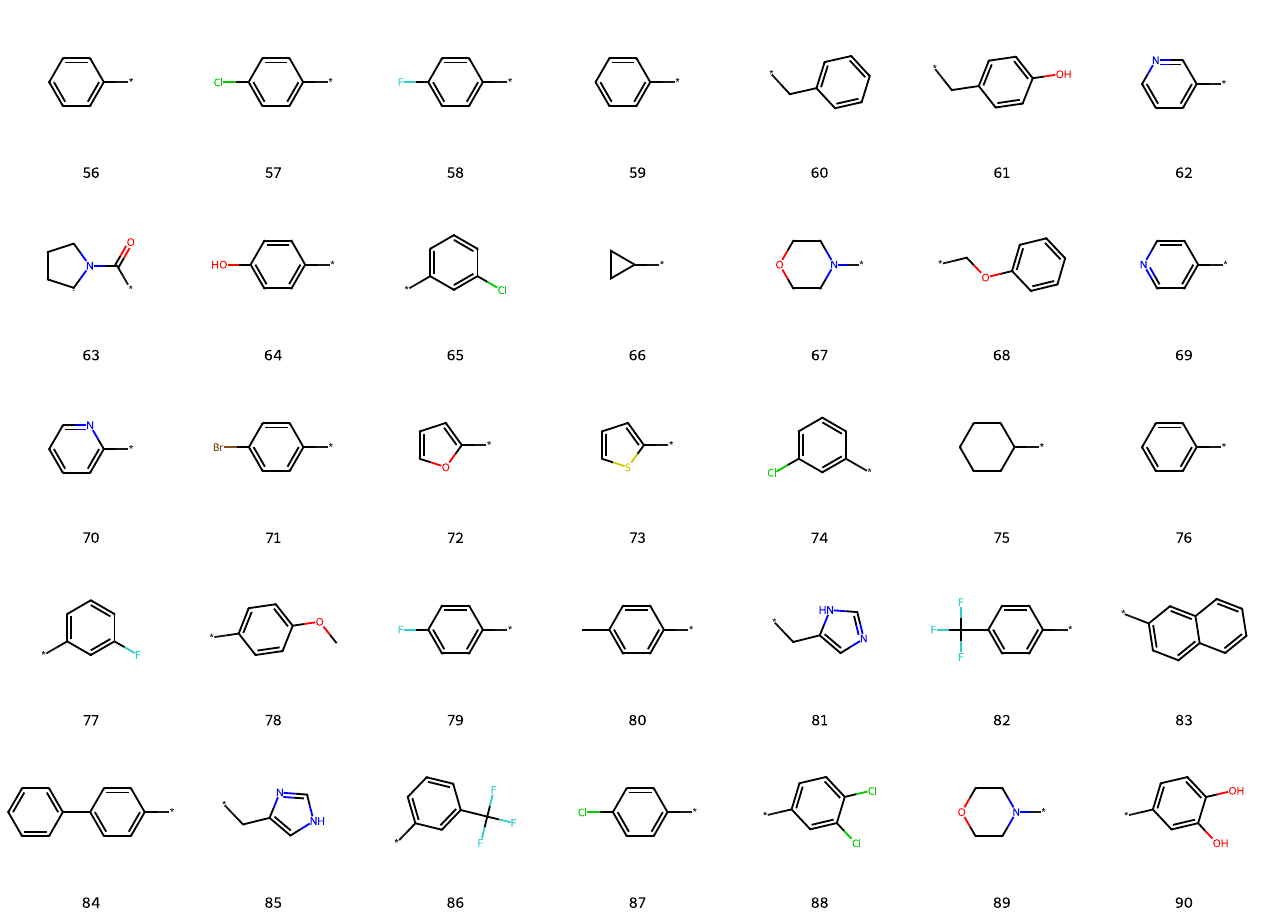}
\caption{Substructures containing complete rings expressed in less than 20 symbols}
\label{fig:small ring groups}
\end{figure}
Most presented rings in Figure \ref{fig:small ring groups} are generally known functional groups. As expected, the functional group benzene is ranked as the most important ring substructure by \textsc{FGCompress}. 
Aside from the substituted benzene derivatives, the rings similarly contain several amino acid side chains (substructures 60 and 61).
The stereochemistry of the disubstituted aromatic groups (substructures 88 and 90) appears reasonable, as the substitutions are positioned as far as possible from the connection point, thereby minimizing steric hindrance. 
\begin{figure}[H]
\centering
   \includegraphics[width=0.7\linewidth]{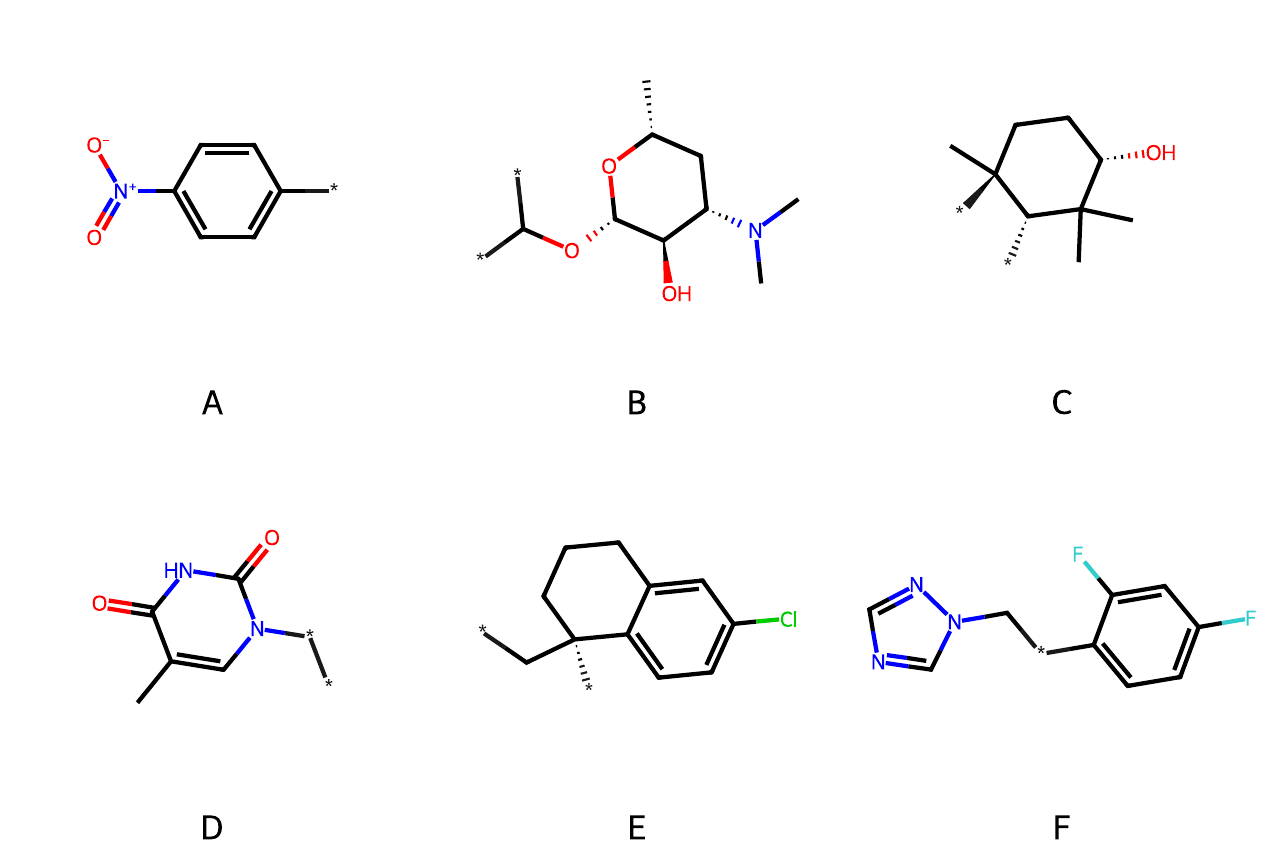}
\caption{Large rings expressed in more than 19 symbols, amide containing ring fragments removed}
\label{fig:large ring groups}
\end{figure}

Larger rings that may be expressed in more than 19 symbols are shown in Figure \ref{fig:large ring groups}. The substructures of Figure \ref{fig:large ring groups} are generally not themselves standard functional groups, but have a more specific biological function. We omit those substructures that are simple fragments of peptides. 
We also omit the discovered substring \texttt{C[C@@H]1OP(=O)(O)OC[C@H]1O[C@@H]}, which, although technically valid, misrepresents the substructure in the original molecules. The original molecules contain two rings, both labelled with 1 to indicate each ring opening and closure. In the extracted substring, the two 1s are misinterpreted as belonging a single ring. We do not observe this specific issue in any other substring.  
Substructure A is a para-nitrobenzene which we do not find surprising. It is common knowledge that nitrobenzenes are ubiquitous in bioactive molecules. 
Substructure B forms part of a desosamine fragment, a central part of the pharmacophore of macrolide antibiotics \cite{burgie2007molecular}. 
Substructure C is found in a number of compounds of the triterpene class such as lanosterol. Lanosterol is the precursor to cholesterol, the compound from which all animal and fungal steroids are derived and plays a role in maintaining lens health \cite{zhao2015lanosterol,nes2011biosynthesis}.
A search on the ChEMBL database reveals that substructure C is present in 169 SMILES strings, and 85 distinct literature sources. These sources refer to investigations into compounds with promise in anticancer, neuroprotective, hepatoprotective, cataract therapy and treatments of Chagas disease \cite{josé2012lanostanoids,nakata2007structure,wang2015design,xu2018hepatoprotective,yang2018synthesis,saidu2024euphane}. 
Substructure C therefore features in a variety of biochemical explanations. 
Substructure D occurs in a number of antiviral therapies, including Zidovudine, Telbivudine and Trifluridine \cite{volberding1990zidovudine, lai20051, carmine1982trifluridine}.
Substructure E features in the MCL-1 inhibitor AMG-176 \cite{caenepeel2018amg}. 
Substructure F mainly occurs in a number of antifungal medications such as Fluconazole and Miconazole \cite{grant1990fluconazole, pierard2012miconazole}.
In addition to those substructures shown in Figure \ref{fig:large ring groups}, the amino acid sequence Leu-Arg-Glu-Phe-Tyr-Gly was also discovered. A search in the ChEMBL database reveals the sequence in peptides which bind Tissue Factor Pathway Inhibitor (TFPI) \cite{tfpi_patent}.

From our analysis we conclude that the answer to \textbf{Q1} is yes, that functional groups emerge from data compression of the ChEMBL dataset. We additionally observe that \textsc{FGCompress} discovers dataset specific functional groups, which correspond to more specific molecular behavior. 

\subsubsection{Q2: Learning Benefit}
Over the 24 datasets, we find that the Ridge Regression models trained using the MML87 fingerprint representation significantly outperform those trained using the MACCS and Morgan fingerprint representations. The performance increase of the MML87 fingerprint was confirmed by Benjamini-Hochberg corrected significance tests at an overall $p<0.05$. 
The MML87 fingerprint outperforms the MACCS in all cases and outperforms the equivalently sized Morgan fingerprint in 17 out of 24 cases. 
The results for each dataset are summarised in table \ref{tab:MML87 vs MACCS}.
\begin{table}[H]
\centering
\caption{Mean Squared Error of the Ridge Regression Models Across Datasets and Fingerprint representations. The standard error from the 5 trials is reported.}
\label{tab:MML87 vs MACCS}
\begin{tabular}{l|ccc}
\textbf{Dataset} & \textbf{MML87} & \textbf{MACCS} & \textbf{Morgan} \\
\hline
B-raf&$0.61\pm0.00$ & $0.72\pm0.01$ & $\boldsymbol{0.49\pm0.01}$ \\
Ephrin&$0.76\pm0.02$ & $0.83\pm0.02$ & $\boldsymbol{0.66\pm0.01}$ \\
Caspase&$\boldsymbol{0.49\pm0.01}$ & $0.67\pm0.00$ & $0.58\pm0.01$ \\
Monoamine&$0.69\pm0.03$ & $0.70\pm0.01$ & $\boldsymbol{0.63\pm0.01}$ \\
Vanilloid&$\boldsymbol{0.61\pm0.01}$ & $0.67\pm0.01$ & $0.67\pm0.00$ \\
Acetylcholinesterase&$\boldsymbol{0.81\pm0.01}$ & $1.11\pm0.01$ & $0.82\pm0.01$ \\
COX-1&$\boldsymbol{0.59\pm0.01}$ & $0.64\pm0.01$ & $0.67\pm0.02$ \\
ABL1&$\boldsymbol{0.81\pm0.00}$ & $0.92\pm0.01$ & $0.93\pm0.01$ \\
A2a&$\boldsymbol{0.63\pm0.03}$ & $1.00\pm0.03$ & $0.80\pm0.02$ \\
COX-2&$\boldsymbol{0.85\pm0.01}$ & $0.93\pm0.01$ & $0.89\pm0.01$ \\
Dihydrofolate&$\boldsymbol{0.90\pm0.03}$ & $0.90\pm0.02$ & $0.92\pm0.02$ \\
opioid&$0.81\pm0.04$ & $0.91\pm0.01$ & $\boldsymbol{0.81\pm0.01}$ \\
Glycogen&$\boldsymbol{0.82\pm0.03}$ & $0.97\pm0.00$ & $0.82\pm0.01$ \\
erbB1&$0.70\pm0.00$ & $1.04\pm0.00$ & $\boldsymbol{0.69\pm0.00}$ \\
LCK&$\boldsymbol{0.80\pm0.00}$ & $1.04\pm0.01$ & $0.95\pm0.00$ \\
Aurora-A&$\boldsymbol{0.88\pm0.00}$ & $1.22\pm0.01$ & $0.88\pm0.01$ \\
Glucocorticoid&$\boldsymbol{0.42\pm0.01}$ & $0.56\pm0.01$ & $0.50\pm0.01$ \\
Cannabinoid&$\boldsymbol{0.61\pm0.01}$ & $0.99\pm0.03$ & $0.73\pm0.02$ \\
Carbonic&$\boldsymbol{0.51\pm0.01}$ & $0.59\pm0.01$ & $0.71\pm0.00$ \\
JAK2&$0.65\pm0.01$ & $0.97\pm0.01$ & $\boldsymbol{0.63\pm0.01}$ \\
HERG&$\boldsymbol{0.41\pm0.00}$ & $0.60\pm0.00$ & $0.56\pm0.00$ \\
Coagulation&$\boldsymbol{0.85\pm0.00}$ & $1.18\pm0.00$ & $1.23\pm0.00$ \\
Estrogen&$\boldsymbol{0.61\pm0.00}$ & $0.68\pm0.00$ & $0.87\pm0.01$ \\
Dopamine&$0.73\pm0.00$ & $0.90\pm0.01$ & $\boldsymbol{0.56\pm0.02}$ \\
    \end{tabular}
\end{table}
These results suggest that the answer to \textbf{Q2} is yes, substructures that compress the dataset are useful for learning and result in improved performance over the MACCS fingerprint on linear regression tasks. 
Note that we do not claim that the MML87 fingerprint is preferable for all learning models, just simple regression. For example, the random forest (RF) model is successful in QSAR tasks and insensitive to large numbers of low information features \cite{olier2018meta, breiman2001random}. Representations that enumerate vast numbers of substructures such as the Morgan fingerprint are instead likely to be beneficial for the RF model. 

\section{Conclusions}
We showed that compressing chemistry results in the automatic identification of the standard set of human-identified functional groups, alongside several dataset specific substructures. This work therefore provides computational validation that partitioning of molecules into functional groups is indeed a good general description of chemistry. We highlighted several interesting patterns that do not form part of the standard chemical functional group set and showed that several compressing functional groups correspond to specific biological function. Empirically, we have shown that fingerprints generated from the discovered substructures aid learning, and that ridge regression models trained on the MML87 fingerprints significantly outperform the MACCS and Morgan  representations on 24 $IC_{50}$ prediction tasks. The improved performance of the MML87 fingerprint over MACCS and Morgan is in spite of the reduced expressiveness of the MML87 fingerprint and comparable fewer substructures compared with the Morgan algorithm. 
\section{Limitations}
For computational tractability, we conducted our search procedure over SMILES strings. Consequently, our search space was restricted to only those substructures which may be represented as SMILES substrings. Unfortunately, SMILES substrings are a subset of the complete set of substructures within a molecule. Searching for subgraphs in graph molecule representations would be complete, but computationally very expensive. Future work should explore a trade-off between the two such as lookahead searches, or hybrid representations. One could adapt recent advances in inductive logic programming (ILP), such as the use of constraint solvers to compress datasets to remedy this \cite{sharma2025learning, cropper2021learning, hocquette2024learning}. 
We demonstrated that compressive substructures enhance predictive performance in learning tasks. However, certain features may be highly informative for explaining specific molecular properties even if they contribute little to overall dataset compression. Moreover, widely used QSAR models such as random forests are relatively insensitive to the inclusion of many low-information features \cite{breiman2001random} and often perform best with representations that enumerate large numbers of substructures, such as the Morgan fingerprint.
The substructures identified by \textsc{FGCompress} therefore remain valuable and could serve as complementary features to enumerative fingerprints, particularly given that \textsc{FGCompress} imposes no restriction on maximum substructure size, unlike many conventional fingerprinting methods.

\providecommand{\latin}[1]{#1}
\makeatletter
\providecommand{\doi}
  {\begingroup\let\do\@makeother\dospecials
  \catcode`\{=1 \catcode`\}=2 \doi@aux}
\providecommand{\doi@aux}[1]{\endgroup\texttt{#1}}
\makeatother
\providecommand*\mcitethebibliography{\thebibliography}
\csname @ifundefined\endcsname{endmcitethebibliography}  {\let\endmcitethebibliography\endthebibliography}{}

\begin{suppinfo}
The complete list of substructures found in the first 500 iterations of the \textsc{FGCompress} algorithm may be found below:
\begin{WrapVerbatim}
(=O)
C(F)(F)F
(C)
C(=O)N
c2ccccc2
c(Cl)c
c1ccccc1
c3ccccc3
c(F)c
NC(=O)
S(=O)(=O)
(O)
C(N)=O
/C=C/
C(=O)O
c([N+](=O)[O-])
CCN
O=C
c(OC)c
c4ccccc4
CCCNC(=N)N
(C#N)
c(C(F)(F)F)c
c(Br)c
(F)
CCCCC
C(=O)
C(C)(C)
S(C)(=O)=O
(N)
NC(=O)[C@H]
C(=O)N[C@@H]
c2ccc(Cl)cc2
CCOCC
(CC(C)C)
C(C)=O
c1ccc(Cl)cc1
COC(=O)
c2ccc(F)cc2
P(=O)(O)O
(CO)
/C=C
/N=C
c1ccc(F)cc1
c3ccc(F)cc3
C(=S)N
(-c2ccccc2)
(-c3ccccc3)
(C(=O)O)
S(N)(=O)=O
[C@@H](O)[C@H]
ccc(O)c
COc
c3ccc(Cl)cc3
(Cc1ccccc1)
cc(Cl)c(Cl)c
cc(OC)c(OC)
[N+](=O)[O-]
(CCCCN)
c5ccccc5
c1-c1c
(CCCNC(=N)N)
O=[N+]([O-])
O=C(O)
/C=N/
c(=O)[nH]
(Cc1ccc(O)cc1)
CC[C@H]
C(C)C
C(=N)N
NC(=O)[C@@H]
c(=O)n
O=S(=O)
S(=O)(=O)N
[C@H](O)
O[C@H](CO)[C@@H](O)[C@H](O)[C@H]
N#C
(CCCN=C(N)N)
CCCC
(CCC(=O)O)
(CC(=O)O)
(OC(C)=O)
CC[C@]
C(F)(F)
C=C
(Cc2ccccc2)
c(N)nc
ccc(OC(F)(F)F)c
C#N
c2cccnc2
CN(C)
(CCC(N)=O)
c(=O)c
C(F)F
(C(F)(F)F)
c(N3CCOCC3)
P(=O)
C(=O)N1CCC[C@H]1
CCN(CC)
N=C(N)
(C(N)=O)
C#C
C(=O)N[C@@H](CCCNC(=N)N)C(=O)N[C@@H]
(N(C)C)
[C@@H](C)
[C@@H](O)
CCN(C)C
c4ccc(F)cc4
c3cccnc3
c2ccc(O)cc2
c1cccnc1
(CC(N)=O)
NC(=O)[C@@H]1CCCN1C(=O)
(C(C)C)
C(C)=C
CC[C@@H]
c2cccc(Cl)c2
C3CC3
ccc([N+](=O)[O-])c
C(=O)N[C@H]
NC(=O)[C@H](CC(C)C)NC(=O)[C@H]
/C=C\
c(-c4ccccc4)
(N2CCOCC2)
/N=C/
O=C(O)C(F)(F)F
c(C(C)(C)C)
c(=O)o
C(=O)N[C@@H](CCCCN)C(=O)N[C@@H]
cc(C#N)c
ccc(C)c
CC(C)
[C@H](C)
c1cccc(Cl)c1
/C=C(\C)
COc1ccccc1
c(N4CCOCC4)
c(F)c(F)c
NS(=O)(=O)
c2ccncc2
c2ccccn2
c2ccc(Br)cc2
(Cl)
c4ccc(Cl)cc4
c1ccco1
c1ccc(Br)cc1
c2cccs2
c2ccco2
c2cc(Cl)ccc2
C2CC2
(=S)
([O-])
/C
ccc(C(F)(F)F)c
C4CC4
c1cccs1
c(C)n
c3cccc(Cl)c3
ncnc
c3ccncc3
C1CCCCC1
CCCCCCCCCC
C2CCCCC2
(c2ccccc2)
C(=O)N[C@@H](CC(C)C)
CCO
ccc(S(N)(=O)=O)c
NC(=O)[C@H](CCCCN)NC(=O)[C@H]
c3ccccn3
(OC)
CC[C@@]
NC(=O)[C@@H](N)C
c1ccc(O)cc1
O[C@H](CO)
c(I)c
c3ccco3
NC(=O)[C@H](CCCNC(=N)N)
c2cccc(F)c2
O[C@@H]
c2ccc(OC)cc2
(-c2ccc(F)cc2)
c1ccccn1
CC[C@H](C)[C@H]
c(O)c
c2ccc(C)cc2
(Cc3ccccc3)
C(=O)N[C@@H](Cc1ccccc1)
c1ccncc1
ccc
[nH]c
c(C)c
Cc1cnc[nH]1
nn(C)c
C[N+](C)(C)
c3ccc(Br)cc3
(NC(C)=O)
cc(Cl)ccc
/C(=N/O)
c3ccc(O)cc3
c2ccc(C(F)(F)F)cc2
c3cccs3
c1cccc(F)c1
NC(=O)OC(C)(C)C
NC(=O)[C@@H]2CCCN2C(=O)
c2ccc3ccccc3c2
c1ccc2ccccc2c1
CC(O)
C3CCCCC3
c3ccc(C)cc3
c1ccc([N+](=O)[O-])cc1
c2ccc([N+](=O)[O-])cc2
/N=N/
(=O)=O
nnc
ccnc
no
c3ccc(OC)cc3
(-c3ccc(F)cc3)
c1ccc(-c2ccccc2)cc1
c3ccc(C(F)(F)F)cc3
/C(C)=C
Cc1c[nH]cn1
CS
c3cccc(F)c3
(S(C)(=O)=O)
c2cccc(C(F)(F)F)c2
(Br)
N=C
(-c2ccc(Cl)cc2)
(C=O)
c1ccc(Cl)c(Cl)c1
c1ccc(C(F)(F)F)cc1
C1CC1
N=[N+]=[N-]
c6ccccc6
ccc(C(=O)O)cc
N1CCOCC1
(c1ccccc1)
cc(OC)c(OC)c(OC)c
csc
c2ccc(-c3ccccc3)cc2
O[C@H](C)C[C@H](N(C)C)[C@H]
[C@@H](O[C@@H]2O[C@H](C)C[C@H](N(C)C)[C@H]2O)
ccc(F)c(Cl)c
c2ccc(Cl)c(Cl)c2
/C(=N\O)
(-c3ccc(Cl)cc3)
C(=O)N[C@@H](CCC(=O)O)
NC(=O)[C@H](Cc1ccccc1)
C(=O)N(C)C
c1ccc(O)c(O)c1
c4cccnc4
[C@H](O)[C@@H]
c1cccc(C(F)(F)F)c1
N(C)C(=O)
NC(=O)[C@H](CCC(=O)O)
c1ccc(OC)cc1
NC(=O)[C@H](CO)
C(=O)N[C@@H](CO)C(=O)N[C@@H]
[C@@H](O)[C@H](O)
[C@@H](C)O
[nH]c(=O)c
[C@@]5(C)CC[C@H](O)C(C)(C)[C@@H]5
(NS(C)(=O)=O)
O[C@H]
C[C@H]
CC(C)C[C@H]
OCO
CCC
C[C@@H]
[C@@H](O)[C@@H]
SCC(=O)N
(Cc2ccc(O)cc2)
(C(=O)OC)
(C)[C@@H]
(-c4cnn(C)c4)
(C(=O)OC(C)(C)C)
c2cc(F)ccc2
C(=O)N[C@@H](C)
C(=O)N[C@@H](CCC(N)=O)
NC(=O)[C@H](CCC(N)=O)
CC(C)(C)
nc(N)c
c3cc(Cl)ccc3
[C@H](O)[C@H]
(C)[C@H]
[S+]([O-])
(C(C)=O)
c3c(N)ncnc3
OC(=O)
O=c
(C3CC3)
Cc1noc(C)c1
c1cc(Cl)ccc1
C(=O)N[C@@H](CC(=O)O)
NC(=O)[C@H](CC(=O)O)
C(=O)N[C@@H](CC(N)=O)
c4ccccn4
(S(=O)(=O)N2CCOCC2)
(-c3ccncc3)
c3cccc(C(F)(F)F)c3
c2ccc(C#N)cc2
NC(=O)[C@H](C)
c(=O)c(C(=O)O)cn
c2nc3ccccc3[nH]2
c1nccs1
[nH]nc
c5ccc(F)cc5
c3ccc(C#N)cc3
C(C)(C)C
c2nccs2
C[C@@H]3OP(=O)(O)OC[C@H]3O[C@@H]
c4ccncc4
c(C(F)(F)F)n
n1-c1
c2ncccc2
c2nc3ccccc3s2
C(=O)N[C@@H](CCCNC(=N)N)
NC(=O)C
NC(=O)c
NC(=O)OC
(OCc2ccccc2)
(CC=C(C)C)
NC(=O)[C@H](CC(N)=O)
CCCCCCCCCCCCCCC
(OC(F)F)c
C(=O)[C@@H]
C(=O)[C@H]
NC(=O)CSc
(c3ccccc3)
N3CCCC3
n4cc(C)c(=O)[nH]c4
c2ccc(F)c(F)c2
ccc(C(C)(C)C)cc
(N3CCCCC3)
c1ccc(C)cc1
(N5CCOCC5)
CCOCCOCCOCCO
(Cc1cnc[nH]1)
(N3CCN(C)CC3)
c1-c1n
CCN1CCOCC1
N2CCCC2
(CC)
(C2CC2)
C(=O)NCC(=O)N[C@@H](Cc1ccc(O)cc1)C(=O)N[C@@H](Cc1ccccc1)C(=O)N[C@@H](CCC(=O)O)C(=O)N[C@@H](CCCNC(=N)N)C(=O)N[C@@H](CC(C)C)
cc(C(F)(F)F)cc(C(F)(F)F)c
c3c(F)cccc3F
ccc(C(=N)N)cc
N1CCCC1
NC(=O)[C@H](CC(C)C)
[C@](C)(O)
NC(=O)[C@H]([C@@H](C)O)
O=P(O)(O)
c3ccc([N+](=O)[O-])cc3
c3ccc(Cl)c(Cl)c3
c2ccnc3cc(Cl)ccc23
(OCc3ccccc3)
c1cccc(C)c1
C(=O)N[C@@H](CCCCN)
cc(OC)c
c4cccc(Cl)c4
c5ccc(Cl)cc5
(C(F)F)
(S)=N
c2nn[nH]n2
(N3CCNCC3)
OCCO
OP(=O)(O)O
CCN2CCOCC2
n2ccnc2
n1ccnc1
c3cc(F)ccc3
cc(Cl)cc(Cl)c
CCN3CCOCC3
B(O)O
(C(O)(C(F)(F)F)C(F)(F)F)
CSSC[C@@H]
c4ccc(O)cc4
C1CCCC1
C2CCCC2
c3ccc4ccccc4c3
CCOC(=O)C
CCOC(=O)c
(CCSC)
(-c2ccncc2)
(-c3ccco3)
[Si](C)(C)
N=C(S)N
C(N)=N
CC(C)=CCC/C(C)=C/C
(-c4ccc(F)cc4)
S(=O)(=O)[O-]
(-c3cccnc3)
N\
ccc(C(F)(C(F)(F)F)C(F)(F)F)c
[C@@](C)
cc(Br)cc
c1cc(F)ccc1
N2CCCCC2
NC(=S)N
(-c3cccs3)
C(=O)N[C@@H](Cc1ccc(O)cc1)
NC(=O)[C@H](Cc1ccc(O)cc1)
c2ccc(O)c(O)c2
CCOP(=O)(OCC)
c2-c2
(N4CCN(C)CC4)
ccc(Cl)c
CC(C)(C)OC(=O)N
cc(F)cc(F)c
ccc(-c2ccccc2-c2nn[nH]n2)c
C5CC5
(-c4ccco4)
(CCCl)CCCl
(-c2ccccc2Cl)
c4cccc(F)c4
N(C)C
c(=O)n(C)
C3CCCC3
ccc(Oc3ccccc3)cc
c1nc2ccccc2[nH]1
(-c2cccs2)
c(Cl)n
N2CCOCC2
n2cnc3c(N)ncnc32
c1nc2ccccc2s1
c2c(Cl)cccc2Cl
COC
C(C)(C)[C@@H]
FC(F)(F)c
(S(=O)(=O)N3CCOCC3)
(-c4cccs4)
ccc(N(C)C)cc
OC(C)=O
(C(=O)[O-])
(CCO)
cc(OC)c(OC)c
NC(=O)[C@@H]2C[C@@H](O)CN2C(=O)[C@@H]
CSSC[C@H]
(c1ccccc1)c1ccccc1
(Cc4ccccc4)
[n+]([O-])c
ccc(S(C)(=O)=O)cc
(n4cc(C)c(=O)[nH]c4=O)
c2c(F)cccc2
/C1=C\c1
C(C)(C)O
/C=C(\C#N)
(-c2cccnc2)
c1ccc(F)c(F)c1
c1c(Cl)cccc1Cl
/C(C#N)=C/
(OS(=O)(=O)O)
(-c3ccccn3)
C(=O)N[C@@H](CO)
(-c3cnn(C)c3)
[C@@]1(CCCc2cc(Cl)ccc21)C
N=C(N)N
N2CCN(C)CC2
P(=O)([O-])
(-c2ccccn2)
cc(F)c(F)c(F)c
c2ccsc2
c1ccsc1
COc1cccc(OC)c1
N1CCCCC1
N3CCOCC3
C3CCOCC3
c1c(F)cccc1
(C4CC4)
c1-c1ccc
C(=O)/C=C/
cc(O)c(O)c(O)c
(-c3ccccc3Cl)
CC(F)(F)C
c(O)cc(O)c
nnn[nH]
c3nc4ccccc4[nH]3
O=C(NO)
O[C@H](CO)[C@H](O)[C@H](O)[C@H]
O[C@H](CO)[C@@H](O)[C@H]
n(C)c
(-c5ccccc5)
c3ncccc3
cc(F)c
n3ccnc3
n1cncn1
(Cn1cncn1)c1ccc(F)cc1F
c3ncnc4[nH]ccc34
c3ccc(-c4ccccc4)cc3
c2ccc(C(=O)O)cc2
c2c(N)ncnc2
OP(=O)(S)
C(=O)N[C@@H](CCCN=C(N)N)
C(=O)N2CCC[C@H]2
(F)(F)(F)(F)F
O=C(Nc1ccccc1)
[C@@H](O)[C@H](O)[C@H]
c3nccs3
\end{WrapVerbatim}

\end{suppinfo}


\end{document}